%% file: jacm-final-cmplg.tex
\documentclass[11pt]{article}
\usepackage{psfig}
\usepackage{fullpage}
\usepackage{eepic}
\usepackage{epsfig}
\usepackage{my-fullname}
\newcounter{treecount}
\newcounter{branchcount}
\newsavebox{\parentbox}
\newsavebox{\treebox}
\newsavebox{\treeboxone}
\newsavebox{\treeboxtwo}
\newsavebox{\treeboxthree}
\newsavebox{\treeboxfour}
\newsavebox{\treeboxfive}
\newsavebox{\treeboxsix}
\newsavebox{\treeboxseven}
\newsavebox{\treeboxeight}
\newsavebox{\treeboxnine}
\newsavebox{\treeboxten}
\newsavebox{\treeboxeleven}
\newsavebox{\treeboxtwelve}
\newsavebox{\treeboxthirteen}
\newsavebox{\treeboxfourteen}
\newsavebox{\treeboxfifteen}
\newsavebox{\treeboxsixteen}
\newsavebox{\treeboxseventeen}
\newsavebox{\treeboxeighteen}
\newsavebox{\treeboxnineteen}
\newsavebox{\treeboxtwenty}
\newlength{\treeoffsetone}
\newlength{\treeoffsettwo}
\newlength{\treeoffsetthree}
\newlength{\treeoffsetfour}
\newlength{\treeoffsetfive}
\newlength{\treeoffsetsix}
\newlength{\treeoffsetseven}
\newlength{\treeoffseteight}
\newlength{\treeoffsetnine}
\newlength{\treeoffsetten}
\newlength{\treeoffseteleven}
\newlength{\treeoffsettwelve}
\newlength{\treeoffsetthirteen}
\newlength{\treeoffsetfourteen}
\newlength{\treeoffsetfifteen}
\newlength{\treeoffsetsixteen}
\newlength{\treeoffsetseventeen}
\newlength{\treeoffseteighteen}
\newlength{\treeoffsetnineteen}
\newlength{\treeoffsettwenty}

\newlength{\treeshiftone}
\newlength{\treeshifttwo}
\newlength{\treeshiftthree}
\newlength{\treeshiftfour}
\newlength{\treeshiftfive}
\newlength{\treeshiftsix}
\newlength{\treeshiftseven}
\newlength{\treeshifteight}
\newlength{\treeshiftnine}
\newlength{\treeshiftten}
\newlength{\treeshifteleven}
\newlength{\treeshifttwelve}
\newlength{\treeshiftthirteen}
\newlength{\treeshiftfourteen}
\newlength{\treeshiftfifteen}
\newlength{\treeshiftsixteen}
\newlength{\treeshiftseventeen}
\newlength{\treeshifteighteen}
\newlength{\treeshiftnineteen}
\newlength{\treeshifttwenty}
\newlength{\treewidthone}
\newlength{\treewidthtwo}
\newlength{\treewidththree}
\newlength{\treewidthfour}
\newlength{\treewidthfive}
\newlength{\treewidthsix}
\newlength{\treewidthseven}
\newlength{\treewidtheight}
\newlength{\treewidthnine}
\newlength{\treewidthten}
\newlength{\treewidtheleven}
\newlength{\treewidthtwelve}
\newlength{\treewidththirteen}
\newlength{\treewidthfourteen}
\newlength{\treewidthfifteen}
\newlength{\treewidthsixteen}
\newlength{\treewidthseventeen}
\newlength{\treewidtheighteen}
\newlength{\treewidthnineteen}
\newlength{\treewidthtwenty}
\newlength{\daughteroffsetone}
\newlength{\daughteroffsettwo}
\newlength{\daughteroffsetthree}
\newlength{\daughteroffsetfour}
\newlength{\branchwidthone}
\newlength{\branchwidthtwo}
\newlength{\branchwidththree}
\newlength{\branchwidthfour}
\newlength{\parentoffset}
\newlength{\treeoffset}
\newlength{\daughteroffset}
\newlength{\branchwidth}
\newlength{\parentwidth}
\newlength{\treewidth}
\newcommand{\ontop}[1]{\begin{tabular}{c}#1\end{tabular}}
\newcommand{\poptree}{%
\ifnum\value{treecount}=0\typeout{QobiTeX warning---Tree stack underflow}\fi%
\addtocounter{treecount}{-1}%
\setlength{\treeoffsettwo}{\treeoffsetthree}%
\setlength{\treeoffsetthree}{\treeoffsetfour}%
\setlength{\treeoffsetfour}{\treeoffsetfive}%
\setlength{\treeoffsetfive}{\treeoffsetsix}%
\setlength{\treeoffsetsix}{\treeoffsetseven}%
\setlength{\treeoffsetseven}{\treeoffseteight}%
\setlength{\treeoffseteight}{\treeoffsetnine}%
\setlength{\treeoffsetnine}{\treeoffsetten}%
\setlength{\treeoffsetten}{\treeoffseteleven}%
\setlength{\treeoffseteleven}{\treeoffsettwelve}%
\setlength{\treeoffsettwelve}{\treeoffsetthirteen}%
\setlength{\treeoffsetthirteen}{\treeoffsetfourteen}%
\setlength{\treeoffsetfourteen}{\treeoffsetfifteen}%
\setlength{\treeoffsetfifteen}{\treeoffsetsixteen}%
\setlength{\treeoffsetsixteen}{\treeoffsetseventeen}%
\setlength{\treeoffsetseventeen}{\treeoffseteighteen}%
\setlength{\treeoffseteighteen}{\treeoffsetnineteen}%
\setlength{\treeoffsetnineteen}{\treeoffsettwenty}%
\setlength{\treeshifttwo}{\treeshiftthree}%
\setlength{\treeshiftthree}{\treeshiftfour}%
\setlength{\treeshiftfour}{\treeshiftfive}%
\setlength{\treeshiftfive}{\treeshiftsix}%
\setlength{\treeshiftsix}{\treeshiftseven}%
\setlength{\treeshiftseven}{\treeshifteight}%
\setlength{\treeshifteight}{\treeshiftnine}%
\setlength{\treeshiftnine}{\treeshiftten}%
\setlength{\treeshiftten}{\treeshifteleven}%
\setlength{\treeshifteleven}{\treeshifttwelve}%
\setlength{\treeshifttwelve}{\treeshiftthirteen}%
\setlength{\treeshiftthirteen}{\treeshiftfourteen}%
\setlength{\treeshiftfourteen}{\treeshiftfifteen}%
\setlength{\treeshiftfifteen}{\treeshiftsixteen}%
\setlength{\treeshiftsixteen}{\treeshiftseventeen}%
\setlength{\treeshiftseventeen}{\treeshifteighteen}%
\setlength{\treeshifteighteen}{\treeshiftnineteen}%
\setlength{\treeshiftnineteen}{\treeshifttwenty}%
\setlength{\treewidthtwo}{\treewidththree}%
\setlength{\treewidththree}{\treewidthfour}%
\setlength{\treewidthfour}{\treewidthfive}%
\setlength{\treewidthfive}{\treewidthsix}%
\setlength{\treewidthsix}{\treewidthseven}%
\setlength{\treewidthseven}{\treewidtheight}%
\setlength{\treewidtheight}{\treewidthnine}%
\setlength{\treewidthnine}{\treewidthten}%
\setlength{\treewidthten}{\treewidtheleven}%
\setlength{\treewidtheleven}{\treewidthtwelve}%
\setlength{\treewidthtwelve}{\treewidththirteen}%
\setlength{\treewidththirteen}{\treewidthfourteen}%
\setlength{\treewidthfourteen}{\treewidthfifteen}%
\setlength{\treewidthfifteen}{\treewidthsixteen}%
\setlength{\treewidthsixteen}{\treewidthseventeen}%
\setlength{\treewidthseventeen}{\treewidtheighteen}%
\setlength{\treewidtheighteen}{\treewidthnineteen}%
\setlength{\treewidthnineteen}{\treewidthtwenty}%
\sbox{\treeboxtwo}{\usebox{\treeboxthree}}%
\sbox{\treeboxthree}{\usebox{\treeboxfour}}%
\sbox{\treeboxfour}{\usebox{\treeboxfive}}%
\sbox{\treeboxfive}{\usebox{\treeboxsix}}%
\sbox{\treeboxsix}{\usebox{\treeboxseven}}%
\sbox{\treeboxseven}{\usebox{\treeboxeight}}%
\sbox{\treeboxeight}{\usebox{\treeboxnine}}%
\sbox{\treeboxnine}{\usebox{\treeboxten}}%
\sbox{\treeboxten}{\usebox{\treeboxeleven}}%
\sbox{\treeboxeleven}{\usebox{\treeboxtwelve}}%
\sbox{\treeboxtwelve}{\usebox{\treeboxthirteen}}%
\sbox{\treeboxthirteen}{\usebox{\treeboxfourteen}}%
\sbox{\treeboxfourteen}{\usebox{\treeboxfifteen}}%
\sbox{\treeboxfifteen}{\usebox{\treeboxsixteen}}%
\sbox{\treeboxsixteen}{\usebox{\treeboxseventeen}}%
\sbox{\treeboxseventeen}{\usebox{\treeboxeighteen}}%
\sbox{\treeboxeighteen}{\usebox{\treeboxnineteen}}%
\sbox{\treeboxnineteen}{\usebox{\treeboxtwenty}}}
\newcommand{\leaf}[1]{%
\ifnum\value{treecount}=20\typeout{QobiTeX warning---Tree stack overflow}\fi%
\addtocounter{treecount}{1}%
\sbox{\treeboxtwenty}{\usebox{\treeboxnineteen}}%
\sbox{\treeboxnineteen}{\usebox{\treeboxeighteen}}%
\sbox{\treeboxeighteen}{\usebox{\treeboxseventeen}}%
\sbox{\treeboxseventeen}{\usebox{\treeboxsixteen}}%
\sbox{\treeboxsixteen}{\usebox{\treeboxfifteen}}%
\sbox{\treeboxfifteen}{\usebox{\treeboxfourteen}}%
\sbox{\treeboxfourteen}{\usebox{\treeboxthirteen}}%
\sbox{\treeboxthirteen}{\usebox{\treeboxtwelve}}%
\sbox{\treeboxtwelve}{\usebox{\treeboxeleven}}%
\sbox{\treeboxeleven}{\usebox{\treeboxten}}%
\sbox{\treeboxten}{\usebox{\treeboxnine}}%
\sbox{\treeboxnine}{\usebox{\treeboxeight}}%
\sbox{\treeboxeight}{\usebox{\treeboxseven}}%
\sbox{\treeboxseven}{\usebox{\treeboxsix}}%
\sbox{\treeboxsix}{\usebox{\treeboxfive}}%
\sbox{\treeboxfive}{\usebox{\treeboxfour}}%
\sbox{\treeboxfour}{\usebox{\treeboxthree}}%
\sbox{\treeboxthree}{\usebox{\treeboxtwo}}%
\sbox{\treeboxtwo}{\usebox{\treeboxone}}%
\sbox{\treeboxone}{\ontop{#1}}%
\sbox{\treeboxone}{\raisebox{-\ht\treeboxone}{\usebox{\treeboxone}}}%
\setlength{\treeoffsettwenty}{\treeoffsetnineteen}%
\setlength{\treeoffsetnineteen}{\treeoffseteighteen}%
\setlength{\treeoffseteighteen}{\treeoffsetseventeen}%
\setlength{\treeoffsetseventeen}{\treeoffsetsixteen}%
\setlength{\treeoffsetsixteen}{\treeoffsetfifteen}%
\setlength{\treeoffsetfifteen}{\treeoffsetfourteen}%
\setlength{\treeoffsetfourteen}{\treeoffsetthirteen}%
\setlength{\treeoffsetthirteen}{\treeoffsettwelve}%
\setlength{\treeoffsettwelve}{\treeoffseteleven}%
\setlength{\treeoffseteleven}{\treeoffsetten}%
\setlength{\treeoffsetten}{\treeoffsetnine}%
\setlength{\treeoffsetnine}{\treeoffseteight}%
\setlength{\treeoffseteight}{\treeoffsetseven}%
\setlength{\treeoffsetseven}{\treeoffsetsix}%
\setlength{\treeoffsetsix}{\treeoffsetfive}%
\setlength{\treeoffsetfive}{\treeoffsetfour}%
\setlength{\treeoffsetfour}{\treeoffsetthree}%
\setlength{\treeoffsetthree}{\treeoffsettwo}%
\setlength{\treeoffsettwo}{\treeoffsetone}%
\setlength{\treeoffsetone}{0.5\wd\treeboxone}%
\setlength{\treeshifttwenty}{\treeshiftnineteen}%
\setlength{\treeshiftnineteen}{\treeshifteighteen}%
\setlength{\treeshifteighteen}{\treeshiftseventeen}%
\setlength{\treeshiftseventeen}{\treeshiftsixteen}%
\setlength{\treeshiftsixteen}{\treeshiftfifteen}%
\setlength{\treeshiftfifteen}{\treeshiftfourteen}%
\setlength{\treeshiftfourteen}{\treeshiftthirteen}%
\setlength{\treeshiftthirteen}{\treeshifttwelve}%
\setlength{\treeshifttwelve}{\treeshifteleven}%
\setlength{\treeshifteleven}{\treeshiftten}%
\setlength{\treeshiftten}{\treeshiftnine}%
\setlength{\treeshiftnine}{\treeshifteight}%
\setlength{\treeshifteight}{\treeshiftseven}%
\setlength{\treeshiftseven}{\treeshiftsix}%
\setlength{\treeshiftsix}{\treeshiftfive}%
\setlength{\treeshiftfive}{\treeshiftfour}%
\setlength{\treeshiftfour}{\treeshiftthree}%
\setlength{\treeshiftthree}{\treeshifttwo}%
\setlength{\treeshifttwo}{\treeshiftone}%
\setlength{\treeshiftone}{0pt}%
\setlength{\treewidthtwenty}{\treewidthnineteen}%
\setlength{\treewidthnineteen}{\treewidtheighteen}%
\setlength{\treewidtheighteen}{\treewidthseventeen}%
\setlength{\treewidthseventeen}{\treewidthsixteen}%
\setlength{\treewidthsixteen}{\treewidthfifteen}%
\setlength{\treewidthfifteen}{\treewidthfourteen}%
\setlength{\treewidthfourteen}{\treewidththirteen}%
\setlength{\treewidththirteen}{\treewidthtwelve}%
\setlength{\treewidthtwelve}{\treewidtheleven}%
\setlength{\treewidtheleven}{\treewidthten}%
\setlength{\treewidthten}{\treewidthnine}%
\setlength{\treewidthnine}{\treewidtheight}%
\setlength{\treewidtheight}{\treewidthseven}%
\setlength{\treewidthseven}{\treewidthsix}%
\setlength{\treewidthsix}{\treewidthfive}%
\setlength{\treewidthfive}{\treewidthfour}%
\setlength{\treewidthfour}{\treewidththree}%
\setlength{\treewidththree}{\treewidthtwo}%
\setlength{\treewidthtwo}{\treewidthone}%
\setlength{\treewidthone}{\wd\treeboxone}}
\newcommand{\branch}[2]{%
\setcounter{branchcount}{#1}%
\ifnum\value{branchcount}=1\sbox{\parentbox}{\ontop{#2}}%
\setlength{\parentoffset}{\treeoffsetone}%
\addtolength{\parentoffset}{-0.5\wd\parentbox}%
\setlength{\daughteroffset}{0in}%
\ifdim\parentoffset<0in%
\setlength{\daughteroffset}{-\parentoffset}%
\setlength{\parentoffset}{0in}\fi%
\setlength{\parentwidth}{\parentoffset}%
\addtolength{\parentwidth}{\wd\parentbox}%
\setlength{\treeoffset}{\daughteroffset}%
\addtolength{\treeoffset}{\treeoffsetone}%
\setlength{\treewidth}{\wd\treeboxone}%
\addtolength{\treewidth}{\daughteroffset}%
\ifdim\treewidth<\parentwidth\setlength{\treewidth}{\parentwidth}\fi%
\sbox{\treebox}{\begin{minipage}{\treewidth}%
\begin{flushleft}%
\hspace*{\parentoffset}\usebox{\parentbox}\\
{\setlength{\unitlength}{2ex}%
\hspace*{\treeoffset}\begin{picture}(0,1)%
\put(0,0){\line(0,1){1}}%
\end{picture}}\\
\vspace{-\baselineskip}
\hspace*{\daughteroffset}%
\raisebox{-\ht\treeboxone}{\usebox{\treeboxone}}%
\end{flushleft}%
\end{minipage}}%
\setlength{\treeoffsetone}{\parentoffset}%
\addtolength{\treeoffsetone}{0.5\wd\parentbox}%
\setlength{\treeshiftone}{0pt}%
\setlength{\treewidthone}{\treewidth}%
\sbox{\treeboxone}{\usebox{\treebox}}%
\else\ifnum\value{branchcount}=2\sbox{\parentbox}{\ontop{#2}}%
\setlength{\branchwidthone}{\treewidthtwo}%
\addtolength{\branchwidthone}{\treeoffsetone}%
\addtolength{\branchwidthone}{-\treeshiftone}%
\addtolength{\branchwidthone}{-\treeoffsettwo}%
\setlength{\branchwidth}{\branchwidthone}%
\setlength{\daughteroffsetone}{\branchwidth}%
\addtolength{\daughteroffsetone}{-\branchwidthone}%
\addtolength{\daughteroffsetone}{-\treeshiftone}%
\setlength{\parentoffset}{-0.5\wd\parentbox}%
\addtolength{\parentoffset}{\treeoffsettwo}%
\addtolength{\parentoffset}{0.5\branchwidth}%
\setlength{\daughteroffset}{0in}%
\ifdim\parentoffset<0in%
\setlength{\daughteroffset}{-\parentoffset}%
\setlength{\parentoffset}{0in}\fi%
\setlength{\parentwidth}{\parentoffset}%
\addtolength{\parentwidth}{\wd\parentbox}%
\setlength{\treeoffset}{\daughteroffset}%
\addtolength{\treeoffset}{\treeoffsettwo}%
\setlength{\treewidth}{\wd\treeboxone}%
\addtolength{\treewidth}{\daughteroffsetone}%
\addtolength{\treewidth}{\treewidthtwo}%
\addtolength{\treewidth}{\daughteroffset}%
\ifdim\treewidth<\parentwidth\setlength{\treewidth}{\parentwidth}\fi%
\sbox{\treebox}{\begin{minipage}{\treewidth}%
\begin{flushleft}%
\hspace*{\parentoffset}\usebox{\parentbox}\\
{\setlength{\unitlength}{0.5\branchwidth}%
\hspace*{\treeoffset}\begin{picture}(2,0.5)%
\put(0,0){\line(2,1){1}}%
\put(2,0){\line(-2,1){1}}%
\end{picture}}\\
\vspace{-\baselineskip}
\hspace*{\daughteroffset}%
\makebox[\treewidthtwo][l]%
{\raisebox{-\ht\treeboxtwo}{\usebox{\treeboxtwo}}}%
\hspace*{\daughteroffsetone}%
\raisebox{-\ht\treeboxone}{\usebox{\treeboxone}}%
\end{flushleft}%
\end{minipage}}%
\setlength{\treeoffsetone}{\parentoffset}%
\addtolength{\treeoffsetone}{0.5\wd\parentbox}%
\setlength{\treeshiftone}{0pt}%
\setlength{\treewidthone}{\treewidth}%
\sbox{\treeboxone}{\usebox{\treebox}}\poptree%
\else\ifnum\value{branchcount}=3\sbox{\parentbox}{\ontop{#2}}%
\setlength{\branchwidthone}{\treewidthtwo}%
\addtolength{\branchwidthone}{\treeoffsetone}%
\addtolength{\branchwidthone}{-\treeshiftone}%
\addtolength{\branchwidthone}{-\treeoffsettwo}%
\setlength{\branchwidthtwo}{\treewidththree}%
\addtolength{\branchwidthtwo}{\treeoffsettwo}%
\addtolength{\branchwidthtwo}{-\treeshifttwo}%
\addtolength{\branchwidthtwo}{-\treeoffsetthree}%
\setlength{\branchwidth}{\branchwidthone}%
\ifdim\branchwidthtwo>\branchwidth%
\setlength{\branchwidth}{\branchwidthtwo}\fi%
\setlength{\daughteroffsetone}{\branchwidth}%
\addtolength{\daughteroffsetone}{-\branchwidthone}%
\addtolength{\daughteroffsetone}{-\treeshiftone}%
\setlength{\daughteroffsettwo}{\branchwidth}%
\addtolength{\daughteroffsettwo}{-\branchwidthtwo}%
\addtolength{\daughteroffsettwo}{-\treeshifttwo}%
\setlength{\parentoffset}{-0.5\wd\parentbox}%
\addtolength{\parentoffset}{\treeoffsetthree}%
\addtolength{\parentoffset}{\branchwidth}%
\setlength{\daughteroffset}{0in}%
\ifdim\parentoffset<0in%
\setlength{\daughteroffset}{-\parentoffset}%
\setlength{\parentoffset}{0in}\fi%
\setlength{\parentwidth}{\parentoffset}%
\addtolength{\parentwidth}{\wd\parentbox}%
\setlength{\treeoffset}{\daughteroffset}%
\addtolength{\treeoffset}{\treeoffsetthree}%
\setlength{\treewidth}{\wd\treeboxone}%
\addtolength{\treewidth}{\daughteroffsetone}%
\addtolength{\treewidth}{\treewidthtwo}%
\addtolength{\treewidth}{\daughteroffsettwo}%
\addtolength{\treewidth}{\treewidththree}%
\addtolength{\treewidth}{\daughteroffset}%
\ifdim\treewidth<\parentwidth\setlength{\treewidth}{\parentwidth}\fi%
\sbox{\treebox}{\begin{minipage}{\treewidth}%
\begin{flushleft}%
\hspace*{\parentoffset}\usebox{\parentbox}\\
{\setlength{\unitlength}{0.5\branchwidth}%
\hspace*{\treeoffset}\begin{picture}(4,1)%
\put(0,0){\line(2,1){2}}%
\put(2,0){\line(0,1){1}}%
\put(4,0){\line(-2,1){2}}%
\end{picture}}\\
\vspace{-\baselineskip}
\hspace*{\daughteroffset}%
\makebox[\treewidththree][l]%
{\raisebox{-\ht\treeboxthree}{\usebox{\treeboxthree}}}%
\hspace*{\daughteroffsettwo}%
\makebox[\treewidthtwo][l]%
{\raisebox{-\ht\treeboxtwo}{\usebox{\treeboxtwo}}}%
\hspace*{\daughteroffsetone}%
\raisebox{-\ht\treeboxone}{\usebox{\treeboxone}}%
\end{flushleft}%
\end{minipage}}%
\setlength{\treeoffsetone}{\parentoffset}%
\addtolength{\treeoffsetone}{0.5\wd\parentbox}%
\setlength{\treeshiftone}{0pt}%
\setlength{\treewidthone}{\treewidth}%
\sbox{\treeboxone}{\usebox{\treebox}}\poptree\poptree%
\else\ifnum\value{branchcount}=4\sbox{\parentbox}{\ontop{#2}}%
\setlength{\branchwidthone}{\treewidthtwo}%
\addtolength{\branchwidthone}{\treeoffsetone}%
\addtolength{\branchwidthone}{-\treeshiftone}%
\addtolength{\branchwidthone}{-\treeoffsettwo}%
\setlength{\branchwidthtwo}{\treewidththree}%
\addtolength{\branchwidthtwo}{\treeoffsettwo}%
\addtolength{\branchwidthtwo}{-\treeshifttwo}%
\addtolength{\branchwidthtwo}{-\treeoffsetthree}%
\setlength{\branchwidththree}{\treewidthfour}%
\addtolength{\branchwidththree}{\treeoffsetthree}%
\addtolength{\branchwidththree}{-\treeshiftthree}%
\addtolength{\branchwidththree}{-\treeoffsetfour}%
\setlength{\branchwidth}{\branchwidthone}%
\ifdim\branchwidthtwo>\branchwidth%
\setlength{\branchwidth}{\branchwidthtwo}\fi%
\ifdim\branchwidththree>\branchwidth%
\setlength{\branchwidth}{\branchwidththree}\fi%
\setlength{\daughteroffsetone}{\branchwidth}%
\addtolength{\daughteroffsetone}{-\branchwidthone}%
\addtolength{\daughteroffsetone}{-\treeshiftone}%
\setlength{\daughteroffsettwo}{\branchwidth}%
\addtolength{\daughteroffsettwo}{-\branchwidthtwo}%
\addtolength{\daughteroffsettwo}{-\treeshifttwo}%
\setlength{\daughteroffsetthree}{\branchwidth}%
\addtolength{\daughteroffsetthree}{-\branchwidththree}%
\addtolength{\daughteroffsetthree}{-\treeshiftthree}%
\setlength{\parentoffset}{-0.5\wd\parentbox}%
\addtolength{\parentoffset}{\treeoffsetfour}%
\addtolength{\parentoffset}{1.5\branchwidth}%
\setlength{\daughteroffset}{0in}%
\ifdim\parentoffset<0in%
\setlength{\daughteroffset}{-\parentoffset}%
\setlength{\parentoffset}{0in}\fi%
\setlength{\parentwidth}{\parentoffset}%
\addtolength{\parentwidth}{\wd\parentbox}%
\setlength{\treeoffset}{\daughteroffset}%
\addtolength{\treeoffset}{\treeoffsetfour}%
\setlength{\treewidth}{\wd\treeboxone}%
\addtolength{\treewidth}{\daughteroffsetone}%
\addtolength{\treewidth}{\treewidthtwo}%
\addtolength{\treewidth}{\daughteroffsettwo}%
\addtolength{\treewidth}{\treewidththree}%
\addtolength{\treewidth}{\daughteroffsetthree}%
\addtolength{\treewidth}{\treewidthfour}%
\addtolength{\treewidth}{\daughteroffset}%
\ifdim\treewidth<\parentwidth\setlength{\treewidth}{\parentwidth}\fi%
\sbox{\treebox}{\begin{minipage}{\treewidth}%
\begin{flushleft}%
\hspace*{\parentoffset}\usebox{\parentbox}\\
{\setlength{\unitlength}{0.5\branchwidth}%
\hspace*{\treeoffset}\begin{picture}(6,1)%
\put(0,0){\line(3,1){3}}%
\put(2,0){\line(1,1){1}}%
\put(4,0){\line(-1,1){1}}%
\put(6,0){\line(-3,1){3}}%
\end{picture}}\\
\vspace{-\baselineskip}
\hspace*{\daughteroffset}%
\makebox[\treewidthfour][l]%
{\raisebox{-\ht\treeboxfour}{\usebox{\treeboxfour}}}%
\hspace*{\daughteroffsetthree}%
\makebox[\treewidththree][l]%
{\raisebox{-\ht\treeboxthree}{\usebox{\treeboxthree}}}%
\hspace*{\daughteroffsettwo}%
\makebox[\treewidthtwo][l]%
{\raisebox{-\ht\treeboxtwo}{\usebox{\treeboxtwo}}}%
\hspace*{\daughteroffsetone}%
\raisebox{-\ht\treeboxone}{\usebox{\treeboxone}}%
\end{flushleft}%
\end{minipage}}%
\setlength{\treeoffsetone}{\parentoffset}%
\addtolength{\treeoffsetone}{0.5\wd\parentbox}%
\setlength{\treeshiftone}{0pt}%
\setlength{\treewidthone}{\treewidth}%
\sbox{\treeboxone}{\usebox{\treebox}}\poptree\poptree\poptree%
\else\ifnum\value{branchcount}=5\sbox{\parentbox}{\ontop{#2}}%
\setlength{\branchwidthone}{\treewidthtwo}%
\addtolength{\branchwidthone}{\treeoffsetone}%
\addtolength{\branchwidthone}{-\treeshiftone}%
\addtolength{\branchwidthone}{-\treeoffsettwo}%
\setlength{\branchwidthtwo}{\treewidththree}%
\addtolength{\branchwidthtwo}{\treeoffsettwo}%
\addtolength{\branchwidthtwo}{-\treeshifttwo}%
\addtolength{\branchwidthtwo}{-\treeoffsetthree}%
\setlength{\branchwidththree}{\treewidthfour}%
\addtolength{\branchwidththree}{\treeoffsetthree}%
\addtolength{\branchwidththree}{-\treeshiftthree}%
\addtolength{\branchwidththree}{-\treeoffsetfour}%
\setlength{\branchwidthfour}{\treewidthfive}%
\addtolength{\branchwidthfour}{\treeoffsetfour}%
\addtolength{\branchwidthfour}{-\treeshiftfour}%
\addtolength{\branchwidthfour}{-\treeoffsetfive}%
\setlength{\branchwidth}{\branchwidthone}%
\ifdim\branchwidthtwo>\branchwidth%
\setlength{\branchwidth}{\branchwidthtwo}\fi%
\ifdim\branchwidththree>\branchwidth%
\setlength{\branchwidth}{\branchwidththree}\fi%
\ifdim\branchwidthfour>\branchwidth%
\setlength{\branchwidth}{\branchwidthfour}\fi%
\setlength{\daughteroffsetone}{\branchwidth}%
\addtolength{\daughteroffsetone}{-\branchwidthone}%
\addtolength{\daughteroffsetone}{-\treeshiftone}%
\setlength{\daughteroffsettwo}{\branchwidth}%
\addtolength{\daughteroffsettwo}{-\branchwidthtwo}%
\addtolength{\daughteroffsettwo}{-\treeshifttwo}%
\setlength{\daughteroffsetthree}{\branchwidth}%
\addtolength{\daughteroffsetthree}{-\branchwidththree}%
\addtolength{\daughteroffsetthree}{-\treeshiftthree}%
\setlength{\daughteroffsetfour}{\branchwidth}%
\addtolength{\daughteroffsetfour}{-\branchwidthfour}%
\addtolength{\daughteroffsetfour}{-\treeshiftfour}%
\setlength{\parentoffset}{-0.5\wd\parentbox}%
\addtolength{\parentoffset}{\treeoffsetfive}%
\addtolength{\parentoffset}{2\branchwidth}%
\setlength{\daughteroffset}{0in}%
\ifdim\parentoffset<0in%
\setlength{\daughteroffset}{-\parentoffset}%
\setlength{\parentoffset}{0in}\fi%
\setlength{\parentwidth}{\parentoffset}%
\addtolength{\parentwidth}{\wd\parentbox}%
\setlength{\treeoffset}{\daughteroffset}%
\addtolength{\treeoffset}{\treeoffsetfive}%
\setlength{\treewidth}{\wd\treeboxone}%
\addtolength{\treewidth}{\daughteroffsetone}%
\addtolength{\treewidth}{\treewidthtwo}%
\addtolength{\treewidth}{\daughteroffsettwo}%
\addtolength{\treewidth}{\treewidththree}%
\addtolength{\treewidth}{\daughteroffsetthree}%
\addtolength{\treewidth}{\treewidthfour}%
\addtolength{\treewidth}{\daughteroffsetfour}%
\addtolength{\treewidth}{\treewidthfive}%
\addtolength{\treewidth}{\daughteroffset}%
\ifdim\treewidth<\parentwidth\setlength{\treewidth}{\parentwidth}\fi%
\sbox{\treebox}{\begin{minipage}{\treewidth}%
\begin{flushleft}%
\hspace*{\parentoffset}\usebox{\parentbox}\\
{\setlength{\unitlength}{0.5\branchwidth}%
\hspace*{\treeoffset}\begin{picture}(8,1)%
\put(0,0){\line(4,1){4}}%
\put(2,0){\line(2,1){2}}%
\put(4,0){\line(0,1){1}}%
\put(6,0){\line(-2,1){2}}%
\put(8,0){\line(-4,1){4}}%
\end{picture}}\\
\vspace{-\baselineskip}
\hspace*{\daughteroffset}%
\makebox[\treewidthfive][l]%
{\raisebox{-\ht\treeboxfour}{\usebox{\treeboxfive}}}%
\hspace*{\daughteroffsetfour}%
\makebox[\treewidthfour][l]%
{\raisebox{-\ht\treeboxfour}{\usebox{\treeboxfour}}}%
\hspace*{\daughteroffsetthree}%
\makebox[\treewidththree][l]%
{\raisebox{-\ht\treeboxthree}{\usebox{\treeboxthree}}}%
\hspace*{\daughteroffsettwo}%
\makebox[\treewidthtwo][l]%
{\raisebox{-\ht\treeboxtwo}{\usebox{\treeboxtwo}}}%
\hspace*{\daughteroffsetone}%
\raisebox{-\ht\treeboxone}{\usebox{\treeboxone}}%
\end{flushleft}%
\end{minipage}}%
\setlength{\treeoffsetone}{\parentoffset}%
\addtolength{\treeoffsetone}{0.5\wd\parentbox}%
\setlength{\treeshiftone}{0pt}%
\setlength{\treewidthone}{\treewidth}%
\sbox{\treeboxone}{\usebox{\treebox}}\poptree\poptree\poptree\poptree%
\else\typeout{QobiTeX warning--- Can't handle #1 branching}\fi\fi\fi\fi\fi}
\newcommand{\tree}{%
\usebox{\treeboxone}
\setlength{\treeoffsetone}{\treeoffsettwo}%
\sbox{\treeboxone}{\usebox{\treeboxtwo}}%
\poptree}

\newcommand{\ab}[1]{ \left \vert #1 \right \vert}

\def\r{\mbox{~$\Rightarrow^*$}~}
\def\W{W}
\def\output{\mbox{${\cal F}_{G,w}$}}
\def\GCNF{\mbox{$G$}}
\def\y{\mbox{``yes''}}
\def\n{\mbox{``no''}}
\def\myeps{\mbox{$\epsilon$}}
\def\gsize{g}
\def\ssize{n}
\newcommand{\slength}{n}

\def\lo{d}

\newcommand{\set}[1]{ \{#1\}}
\newcommand{\qset}[2]{ \{{#1} : {#2}\}}
\newcommand{\tg}[1]{T(#1)}
\newcommand{\tw}[1]{t(#1)}

\newcommand{\cpa}[4]{ 
\begin{equation}
\renewcommand{\theequation}{#1}~~~~~~~~
{#2} \longrightarrow {#3},~~ {#4}
\end{equation}
}

\newcommand{\proof}[1]{
{\noindent {\it Proof.} {#1} \rule{2mm}{2mm} \vskip \belowdisplayskip}
}

\newtheorem{definition}{Definition}

\newtheorem{claim}{Claim}
\newtheorem{corollary}{Corollary}
\newtheorem{theorem}{Theorem}

\newcommand{\epsfscaledbox}[2]{\centerline{\psfig{figure=#1,width=#2}}}

\author{Lillian Lee \\  Department of Computer Science, Cornell University}

\title{\vspace{-55pt}
{\normalsize \tt \hfill To appear in {\em Journal of the ACM}, 2002} \\
        \mbox{}\\Fast Context-Free Grammar Parsing Requires Fast Boolean Matrix
Multiplication} 
\date{}

\begin{document}
\bibliographystyle{my-fullname}

\maketitle

\begin{abstract}
In 1975, Valiant showed that Boolean matrix multiplication can be used
for parsing context-free grammars (CFGs), yielding the asympotically
fastest (although not practical) CFG parsing algorithm known.  We
prove a dual result: any CFG parser with time complexity
$O(g\slength^{3 - \myeps})$, where $g$ is the size of the grammar and
$\slength$ is the length of the input string, can be efficiently
converted into an algorithm to multiply $m \times m$ Boolean matrices
in time $O(m^{3 - \myeps/3})$.
Given that practical, substantially sub-cubic Boolean matrix
multiplication algorithms have been quite difficult to find, we thus
explain why there has been little progress in developing practical,
substantially sub-cubic general CFG parsers.  In proving this result,
we also develop a formalization of the notion of parsing.
\end{abstract}

\section{Introduction}

The context-free grammar (CFG) formalism, introduced by
\namecite{Chomsky:56}, has enjoyed wide use in a variety of fields.
CFGs have been used to model the structure of programming languages,
human languages, and even biological data such as the sequences of
nucleotides making up DNA and RNA
\cite{Aho+Sethi+Ullman:86a,Jurafsky+Martin:00a,Durbin+Eddy+al:98a}.

CFGs are generative systems, where strings are derived
via successive applications of rewriting rules.  In practice, however,
the goal generally is not to generate valid strings from a grammar.
Rather, one typically already has some string of interest, such as a C
program or an English sentence, in hand, and the goal is to analyze
--- {\em parse} --- the string with respect to the grammar.

Canonical methods for general CFG parsing are the CKY algorithm
\cite{Kasami:cky,Younger:cky} and Earley's algorithm
\cite{Earley:parse}. 
Both have a worst-case running time of $O(\gsize\ssize^3)$ for a CFG
of size $\gsize$ and string of length $\ssize$
\cite{Graham+Harrison+Ruzzo:journal}, although CKY requires the input
grammar to be in Chomsky normal form in order to achieve this time bound.
Unfortunately, cubic dependence on the string length is prohibitively
expensive in applications such as speech recognition, where responses
must be made in real time, or in situations where the input sequences
are very long, as in computational biology.

Asymptotically faster parsing algorithms do exist.
\namecite{Graham+Harrison+Ruzzo:journal} give a variant of Earley's
algorithm that is based on the so-called ``four Russians'' algorithm
\cite{four-russians} for Boolean matrix multiplication (BMM); it runs
in time $O(\gsize\ssize^3/\log \ssize)$.  \namecite{Rytter:85a}
further modifies this parser by a compression technique, improving the
dependence on the string length to $O(\ssize^3/\log^2 \ssize)$. But
Valiant's \shortcite{Valiant:cfl} parsing method, which reorganizes
the computations of CKY, is the asymptotically fastest known.  It also
uses BMM; its worst-case running time for a grammar in Chomsky normal
form is proportional to $M(\ssize)$, where $M(m)$ is the time it takes to
multiply two $m \times m$ Boolean matrices together.

\begin{figure}
\epsfscaledbox{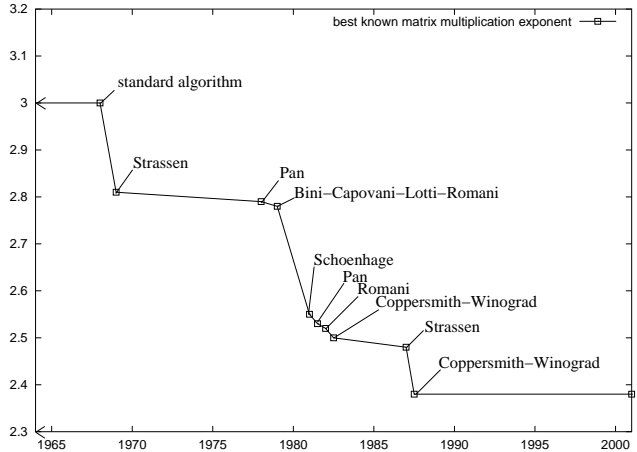}{3.5in}
\caption{\label{fig:multexp} Lowest known upper bound on the exponent
$\omega$ for the complexity of matrix multiplication.  For instance,
before 1969, the fastest known algorithm for matrix multiplication
took proportional to $m^3$ steps ($\omega=3$).}
\end{figure}

Since these subcubic parsing algorithms all depend on Boolean matrix
multiplication, it is natural to ask how fast BMM can be performed in
practice.  The asymptotically fastest way known to perform BMM is to
rely on algorithms for multiplying arbitrary matrices.  There exist
matrix multiplication algorithms with time complexity $O(m^{3-
\delta})$, thus improving over the standard algorithm's $O(m^3)$
running time; for instance, Strassen's \shortcite{Strassen:mult} has a
worst-case running time of $O(m^{2.81})$, and the fastest currently
known, due to Coppersmith and Winograd (1987;1990), has time
complexity $O(m^{2.376})$.  (See \namecite{Strassen:hist} for a
historical account, plotted graphically in figure \ref{fig:multexp}.)
\nocite{Coppersmith+Winograd:87a,Coppersmith+Winograd:90a}
Unfortunately, the constants involved in the subcubic
algorithms improving on Strassen's result are so large that these {\em
fast} algorithms cannot be used in practice.  As for Strassen's
method itself, its practicality is ambiguous: empirical studies show that the
``cross-over'' point --- the matrix size at which it becomes better to
use Strassen's method ---  is above 100
\cite{Bailey:88a,Thottethodi+Chatterjee+Lebeck:98a}.  In summary,
despite decades of research effort, there has been little success
at finding a clearly practical, simple, fast matrix multiplication
algorithm.

One might therefore hope to find a way to speed up CFG parsing without
relying on matrix multiplication.  However, the main theorem of this
paper is that fast CFG parsing {\em requires} fast Boolean matrix
multiplication, in the following precise sense: any parser running in
time $O(g\slength^{3 - \myeps})$ that represents parse data in a
retrieval-efficient way can be converted with little computational
overhead into an $O(m^{3 - \myeps / 3})$ BMM algorithm.  

The restriction of our result to parsers with a linear dependence on
the grammar size is crucial for relating sub-cubic parsing to
sub-cubic BMM.  However, as discussed in section
\ref{sec:analyze-runtimes}, this restriction is a reasonable one since
canonical parsing algorithms such as CKY and Earley's algorithm have
this property, and furthermore, in domains like natural language
processing, the grammar size is often the dominating factor.

Our theorem, together with the fact that it has been quite difficult
to find practical fast matrix multiplication algorithms,
explains why there has been little success to date in developing practical CFG
parsers running in substantially sub-cubic time.

\section{The parsing problem: a formalization}

In this section, we motivate and set forth a formalization of the
parsing problem.  

\subsection{Motivation for our definition}

In formal language theory, emphasis has been placed on the {\em
recognition} or membership problem: deciding whether or not a given
string can be derived by a grammar.  However, we concentrate here on
the {\em parsing} problem: finding the parse structure, or analysis,
assigned to a string by a grammar. (In the case of {\em ambiguous}
strings, multiple parses exist; we address this point below.)

From a theoretical standpoint, the two problems are almost
equivalent. Recognition obviously reduces to parsing, and indeed to
our knowledge there are no CFG recognition algorithms that do not 
implicitly compute parse information.
Conversely, \namecite{Ruzzo:79a} demonstrated that any CFG recognition
algorithm that is not already an implicit parser can be converted
into an algorithm that returns a (single) parse of the input string
$w$, at a cost of only a factor of $O(\log \ab{w})$ slowdown.

\begin{figure}
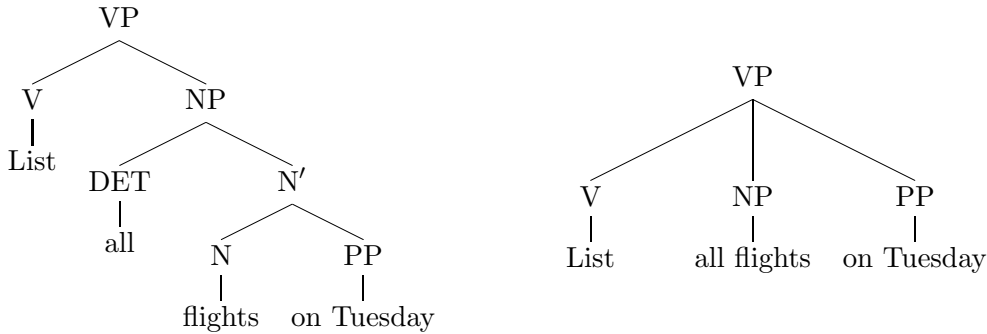

\begin{tabular}{cc}
\leaf{List}
\branch{1}{$\mbox{\rm V}$}
\leaf{all}
\branch{1}{$\mbox{\rm DET}$}
\leaf{flights}
\branch{1}{$\mbox{\rm N}$}
\leaf{on Tuesday}
\branch{1}{$\mbox{\rm PP}$}
\branch{2}{$\mbox{\rm N}'$}
\branch{2}{$\mbox{\rm NP}$}
\branch{2}{$\mbox{\rm VP}$}
\tree
&
\leaf{List}
\branch{1}{$\mbox{\rm V}$}
\leaf{all flights}
\branch{1}{$\mbox{\rm NP}$}
\leaf{on Tuesday}
\branch{1}{$\mbox{\rm PP}$}
\branch{3}{$\mbox{VP}$}
\tree 
\end{tabular}
\caption{\label{fig:flights} Two different parse trees
for the sentence ``List all flights on Tuesday''. The labels on the
interior nodes denote linguistic categories.}
\end{figure}

In practice, however, the parsing problem is much more compelling than
the membership problem. Understanding the structure of the input
string is crucial to programming language compilation, natural
language understanding, RNA shape determination, and so
on. In fact, in speech recognition systems, a useful assumption
is that any input utterance is somehow ``valid'', even if it is
ungrammatical,  thus making the recognition problem trivial.  However,
different
parses of the input sentence may lead to radically different
interpretations.  For example, the classic sentence ``List all flights
on Tuesday'' has two different parses (see Figure \ref{fig:flights}):
one indicates that all flights taking off on Tuesday should be listed
right now, whereas the other asks to wait until Tuesday, and then list
all flights regardless of their departure date.  Another well-known
ambiguous sentence is ``I saw the man with the telescope''; observe
that here the two possible interpretations seem to be about equally
likely.

The fact that some input strings are ambiguous raises the question of
what we should require the output of a parsing algorithm to be: any
{\em single} parse of the input string (Ruzzo's reduction of parsing
to recognition uses this model), or {\em all
possible} parses? In practice, since multiple analyses may be valid
(as in the natural language examples above), it is clear that any
practical parser should return all parses.

It remains to determine what the format of the output parses should
be.  One problem is that there exist grammars in which the number of
parse trees for strings of length $n$ grows exponentially in $n$; for
example, consider the Chomsky normal form CFG with productions $S
\rightarrow SS | a$.\footnote{If we do not impose any restrictions on
the form of the grammar, then an {\em infinite} number of parse trees
can be produced for a single string; for example, consider the
production set $S \rightarrow S|a$.}
Hence, a compressed
representation of the parse structures must be used; otherwise, every
parser could take exponential time just to print its output.
However, we must be careful to impose restrictions on the compression
rate: after all, we could perversely consider the input string itself
to be a (rather inconvenient) representation of all its parse trees
\cite{Ruzzo:79a}.  
We thus require practical parsers to output all the parses of an input
string in a representation that is both compact and yet allows
efficient retrieval of parse information.  
In the next subsection, we make this notion precise.

\subsection{C-parsing of context-free grammars}

We use the usual definition of a context-free grammar (CFG) as a
4-tuple $G=(\Sigma, V, R, S)$, where $\Sigma$ is the set of terminals,
$V$ is the set of nonterminals, $R$ is the set of rewrite rules or
productions, and $S \in V$ is the start symbol.  Given a string $w =
w_1 w_2 \cdots w_\slength$ in $\Sigma^*$, where each $w_i$ is an element of
$\Sigma$, we use the notation $w_i^j$ to denote the substring $w_i
w_{i+1} \cdots w_{j-1} w_j$.  
The {\em size} of $G$, denoted by $|G|$, is the sum of the lengths of
all productions in $R$.

Our notion of necessary parse information is based on the concept of CFG
{\em c-derivations}, which are substring derivations that are
consistent with some parse of the entire input string.
\begin{definition}
Let $G=(\Sigma, V, R, S)$ be a CFG, and let $w= w_1 w_2 \cdots w_\slength$,
$w_i \in \Sigma$.  A nonterminal
$A \in V$ {\em c-derives} (consistently derives) $w_i^j$ if and only
if the following conditions hold:
\begin{itemize}
\item $A \r w_i^j$, and
\item $S \r w_1^{i-1} A w_{j+1}^\slength.$
\end{itemize}
(These conditions together imply that $S \r w$.)  
\end{definition}

We argue, as do \namecite{Ruzzo:79a} and, for a different formalism,
\namecite{Satta:bmm}, that a practical parser must create output from
which c-derivation information can be retrieved efficiently.  This
information is what allows us to ascertain that there exists an
analysis of the input sequence for which a certain substring forms a
{\em constituent}, or coherent unit.  In contrast, derivation
information records potential subderivations that may not be
consistent with any analysis of the full input string.  For example,
in the sentence ``Only the lonely can play'', ``the lonely can'' could
conceivably, in isolation, form a noun phrase, but clearly in any
reasonable grammar of English no nonterminal c-derives that substring.
While some parsers retain information about derivations that are not
c-derivations, we formulate our definition of parsing to include
algorithms that do not.
\begin{definition}
A {\em c-parser} is an algorithm that takes a CFG  $G=(\Sigma,
V, R, S)$ and string $w \in \Sigma^*$ as input and
produces output $\output$ that acts as an oracle about parse
information as follows: for any $A \in V$, 
\begin{itemize}
\item If $A$ c-derives $w_i^j$, then $\output(A,i,j) = \y$.
\item If $A \not \Rightarrow^* w_i^j$ (which implies that $A$ does not
c-derive $w_i^j$), then $\output(A,i,j) = \n.$
\item $\output$ answers queries in constant time.
\end{itemize}
\end{definition}

The asymmetry of derivation and c-derivation in our definition of
c-parsing is deliberate.  We allow $\output$'s answer to be arbitrary
if $A \r w_i^j$ but $A$ does not c-derive $w_i^j$; we leave it to the
algorithm designer to decide which answer is appropriate.  Thus, our
definition makes the class of c-parsers as broad as possible: if we
had changed the first condition to ``If $A$ derives $w_i^j \ldots$'',
then Earley parsers would be excluded, since they do not keep track of
all substring derivations; whereas if we had written the second
condition as ``If $A$ does not c-derive $w_i^j$, \ldots'', then
CKY would not be a c-parser, since it tracks all substring
derivations, not just c-derivations.  In fact, the class of c-parsers
contains all {\em tabular} parsers, including generalized LR parsing,
CKY, and Earley's algorithm \cite{Nederhof+Satta:96a}.  In contrast,
\namecite{Ruzzo:79a} deals with the difference between derivations and
c-derivations by defining two different problems (the {\em weak all-parses
problem} and the {\em all-parses problem}).

Our choice of an oracle rather than a specific data structure as the
output of a c-parser is also for the purpose of keeping our definition
as broad as possible.  In tabular algorithms like CKY, the oracle is
given in the form of a matrix or {\em chart}; indeed, Ruzzo's
\shortcite{Ruzzo:79a} definition of the all-parses and weak all-parses
problems requires the output to be a matrix.  However (as Ruzzo points
out), this is not the only possibility, and furthermore has a
liability from a technical point of view: if the output must be a
matrix, then all parsing algorithms must take time at least
$\Omega(\ssize^2)$ even to print their output.  Since it may be
possible for c-derivations to be represented more compactly, we prefer
to allow for this possibility in our definition.

Finally, with regards to the third condition, we observe that
\namecite{Satta:bmm} imposes the same constant-time constraint for a
different grammar formalism (tree-adjoining grammars).
On the other
hand, we could loosen this to allow query processing to take time
polylogarithmic in the string and grammar size without much effect on
our results (see section \ref{sec:timebounds}).

\subsection{Analyzing parser runtimes}
\label{sec:analyze-runtimes}

It is common in the formal language theory literature to see the
running time of parsing algorithms described as a function of the
length of the input string only (e.g., $O(\slength^3)$ for a string of
length $\slength$).  That is, the size of the context-free grammar is
often treated as a constant.  This stems in part from two
characteristics of the programming languages and compilers domains:
first, the size of a computer program's source code is typically much
greater than the size of the grammar describing the programming
language's syntax, so that the grammar term is negligible; and second,
compilers are constructed to analyze many different programs with
respect to a single built-in grammar.

However, in other domains these conditions do not hold.  For example,
in natural language, sentences are relatively short (not often longer
than one hundred words) compared with the size of the grammar:
\namecite{Johnson:98a} describes a (probabilistic) CFG for a subset
of English that has 22,773 rules.  
Indeed, \namecite{Joshi:97a} notes that ``the real limiting factor in
practice is the size of the grammar''.
Therefore, it is reasonable to include in the analysis of parsing time
the dependence on the grammar size, and we will do so here.  As
a point of information, we note that both CKY and Earley's algorithm
can be implemented to run in time $O(|G|n^3)$
\cite{Graham+Harrison+Ruzzo:journal}, although CKY requires the input
grammar to be in Chomsky normal form, conversion to which may cause a
quadratic increase in the number of productions in the grammar
\cite{Hopcroft:79a}.

\section{The reduction}

In this section, we provide two efficient reductions of Boolean matrix
multiplication to c-parsing, thus proving that any c-parsing algorithm
can be used as a Boolean matrix multiplication algorithm with little
computational overhead.  The first reduction produces a string and a
context-free grammar; the second is a modification of the first in
which the grammar produced is in Chomsky normal form.  The techniques
we use are an adaptation of Satta's \shortcite{Satta:bmm} elegant
reduction of Boolean matrix multiplication to {\em tree-adjoining
grammar} (TAG) parsing.  However, Satta's results rely explicitly on
properties of TAGs that allow them to generate non-context-free
languages, and so cannot be directly applied to CFGs.

\subsection{Boolean matrix multiplication}

A Boolean matrix is a matrix with entries from the set $\set{0,1}$.  A
Boolean matrix multiplication (BMM) algorithm takes as input two $m
\times m$ Boolean matrices $A$ and $B$ and returns their {\em Boolean
product} $A \times B$, which is the $m \times m$ Boolean matrix $C$
whose entries are defined by
$$c_{ij} = \bigvee_{k=1}^m \left( a_{ik} \wedge b_{kj} \right).$$ That
is, $c_{ij} =1$ if and only if there exists a number $k$, $1 \leq k
\leq m$, such that $a_{ik} = b_{kj} = 1$.

As noted above, the Boolean product $C$ can be computed via standard matrix
multiplication, since $c_{ij} = \sum_{k=1}^m a_{ik} \cdot b_{kj}$.
This means that we can use the \namecite{Coppersmith+Winograd:90a}
general matrix multiplication algorithm to calculate the Boolean
matrix product of two $m \times m$ Boolean matrices in time
$O(m^{2.376})$.  To our knowledge, the asympotically fastest
algorithms for BMM all rely on general matrix multiplication; the
fastest algorithms that do not do so are the so-called ``four
Russians'' algorithm \cite{four-russians}, with worst-case running time
$O(m^3/\log(m))$, and Rytter's \shortcite{Rytter:85a} variant which
uses compression to reduce the time to $O(m^3/\log^2(m))$.

\subsection{The reduction: first version}

\begin{figure}
\epsfscaledbox{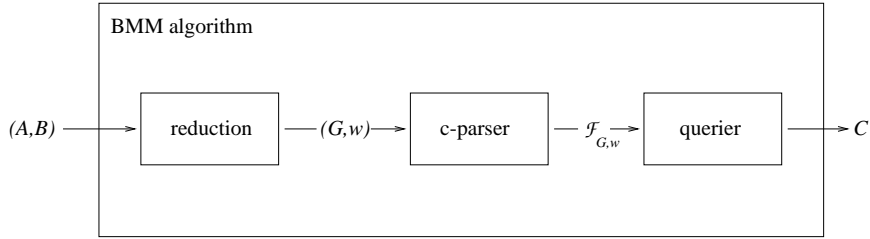}{4.5in}
\caption{\label{fig:createBMM} Converting a c-parser into a BMM algorithm.}
\end{figure}

Our goal in this section is to show that Boolean matrix multiplication
can be efficiently reduced to c-parsing of CFGs.  That is, we will
describe a simple procedure that takes as input an instance of the BMM
problem and converts it into an instance of the CFG parsing problem
with the following property: any c-parsing algorithm run on the new
parsing problem yields output from which it is easy to determine
the answer to the original BMM problem.  We therefore demonstrate that
any c-parser can be used to solve Boolean matrix
multiplication via the three-step process shown schematically in
Figure \ref{fig:createBMM}.

Thus, given two Boolean matrices $A$ and $B$, we need show how to produce a
grammar $G$ and a string $w$ such that c-parsing $w$ with respect to
$G$ yields output $\output$ from which information about the Boolean
product $C = A \times B$ can be easily retrieved.  Our approach will
be to encode almost all the information about $A$ and $B$ in the
grammar.

We can sketch the desired behavior of the grammar $G$ as follows.
Suppose entries $a_{ik}$ in $A$ and $b_{kj}$ in $B$ are both 1.
Assume we have some way to break up array indices into two parts so
that $i$ can be reconstructed from $i_1$ and $i_2$, $j$ can be
reconstructed from $j_1$ and $j_2$, and $k$ can be reconstructed from
$k_1$ and $k_2$ (we will describe a way to do this later; the
motivation is to keep the grammar size relatively small).
Then, our grammar will permit the following derivation sequence:

\begin{eqnarray*}
C_{i_1,j_1} &\Rightarrow \: & A_{i_1,k_1} B_{k_1,j_1} \\
 &\Rightarrow^*& 
\underbrace{w_{i_2}  \cdots w_{k_2 + \delta}}_{\mbox{derived by $A_{i_1,k_1}$}} 
\underbrace{w_{k_2 + \delta + 1} \cdots w_{j_2 + 2
\delta}}_{\mbox{derived by $B_{k_1,j_1}$}} ,
\end{eqnarray*}
where $\delta$ will be defined later.
The key thing to observe is that $C_{i_1,j_1}$ generates two
nonterminals whose ``inner'' indices match, and that these two
nonterminals generate substrings that lie exactly next to each other.
The ``inner'' indices constitute a check on $k_1$, and substring
adjacency constitutes a check on $k_2$; together, these two checks
serve as a proof that $a_{ik}=b_{kj}=1$, and hence that
$c_{ij}$ is also 1.  

We now set up some notation.  Let $A$ and $B$ be two Boolean matrices,
each of size $m \times m$, and let $C$ be their Boolean matrix
product.  In the rest of this section, we consider $A$, $B$, $C$, and
$m$ to be fixed.  Set $\lo = \lceil m^{1/3} \rceil$, and set $\delta =
\lo +2$. (The effect of these choices on the efficiency of our
reduction is discussed in section \ref{sec:timebounds}.)  We will be
constructing a string of length $3 \delta$; we choose $\delta$
slightly larger than $\lo$ in order to avoid having
epsilon-productions in our grammar.

Our index encoding function is as follows.  Let $i$ be a matrix
index, $1 \leq i \leq m \leq \lo^3$.  Then, we define the function $f(i)=
(f_1(i), f_2(i))$ by
\begin{eqnarray*}
f_1(i) & = & \lfloor i/\lo \rfloor ~~~~~~~~~~~~~  (\mbox{so that } 0 \leq f_1(i) \leq
\lo^2), ~ \mbox{and} \\
f_2(i) & = & (i ~ {\rm mod} ~ \lo) + 2 ~~ (\mbox{so that } 2 \leq f_2(i) \leq \lo+1).
\end{eqnarray*}
Since $f_1(i)$ and $f_2(i)$ are essentially the quotient and remainder
of integer division of $i$ by $\lo$, we can reconstruct $i$ from
$(f_1(i), f_2(i))$.  It may be helpful to think of these two
quantities as ``high-order'' and ``low-order'' bits, respectively. For
convenience, we will employ the notational shorthand of using
subscripts instead of the functions $f_1$ and $f_2$; that is, we write
$i_1$ and $i_2$ for $f_1(i)$ and $f_2(i)$.

It is now our job to create a CFG $G = (\Sigma,V,R,S)$ and a string
$w \in \Sigma^*$ that encode information about $A$ and $B$ and express
constraints about their product $C$.

We choose the set of terminals to be $\Sigma = \qset{w_\ell}{1 \leq
\ell \leq 3\lo + 6}$.  The string we choose is extremely simple, and in
fact doesn't depend on $A$ or $B$ at all: we set $w = w_1 w_2 \cdots
w_{3\lo+6}$.  We consider $w$ to be made up of three parts, $x$, $y$,
and $z$, each of size $\delta$: $$w = \underbrace{w_1 w_2 \cdots
w_{\lo+2}}_{x} \underbrace{w_{\lo+3} \cdots w_{2\lo+4}}_{y}
\underbrace{w_{2\lo+5} \cdots w_{3\lo+6}}_z .$$ Observe that for any array
index $i$ between $1$ and $m$, it is the case that $w_{i_2}$ appears
in $x$, $w_{i_2 + \delta}$ appears in $y$, and $w_{i_2 + 2 \delta}$
appears in $z$, since
\begin{eqnarray*}
 i_2 & \in & [2, \lo+1], \\
 i_2 + \delta & \in & [\lo+4, 2\lo+3], \mbox{ and} \\
i_2 + 2 \delta & \in & [2\lo + 6, 3\lo+5].
\end{eqnarray*}

We now turn our attention to constructing the grammar $G$.  Our plan
is to include a set of nonterminals $\qset{C_{p,q}}{1 \leq p, q \leq
\lo^2}$ in $V$ such that $c_{ij} = 1$ if and only if $C_{i_1,j_1}$
c-derives $w_{i_2}^{j_2 + 2 \delta}$.

\subsection{The grammar}
\label{grammar1}
To create $G=(\Sigma, V, R, S)$,  we build up the set of  nonterminals
and productions, starting with $V =
\set{S}$ and  $R = \emptyset$.  We add nonterminal
$\W$ to $V$ for generating arbitrary non-empty substrings and
therefore add productions \cpa{$\W$-\mbox{rules}}{\W}{w_\ell \W \vert
w_\ell}{1 \leq \ell \leq 3\lo+6.}  

Next, we encode the entries of the
input matrices $A$ and $B$ in our grammar.  We add the
nonterminals from the sets $\qset{A_{p,q}}{1 \leq p, q \leq \lo^2}$ and
$\qset{B_{p,q}}{1 \leq p, q \leq \lo^2}$.  Then, for every {\em
non-zero} entry $a_{ij}$ in $A$, we add the production
\begin{equation}
\renewcommand{\theequation}{$A$-\mbox{rules}} 
{A_{i_1,j_i}} \longrightarrow {w_{i_2} \W w_{j_2 + \delta}.}
\label{A-rules}
\end{equation}
For every {\em non-zero} entry $b_{ij}$ in $B$, we add the production
\begin{equation}
\renewcommand{\theequation}{$B$-\mbox{rules}} 
~~~~~~{B_{i_1,j_i}} \longrightarrow {w_{i_2 + 1 + \delta} \W w_{j_2 + 2 \delta}.}
\label{B-rules}
\end{equation}
To represent the entries of $C$, we
add the nonterminals from the set $\qset{C_{p,q}}{1 \leq p, q \leq
\lo^2}$ and include  productions
\begin{equation}
\renewcommand{\theequation}{$C$-\mbox{rules}}
\hspace{.6in}{C_{p,q}} \longrightarrow {A_{p,r} B_{r,q}},~~ {1 \leq p,
q, r \leq \lo^2.}
\label{C-rules}
\end{equation}
Finally, we complete the construction with productions for the
start symbol $S$:
\cpa{$S$-\mbox{rules}}{S}{\W C_{p,q} \W}{1 \leq p, q \leq \lo^2.}

We now prove the following result about the grammar and string
we have just described.
\begin{theorem}
For $1 \leq i, j \leq m$, the entry 
$c_{ij}$ in $C$ is non-zero if and only if $C_{i_1,j_1}$
c-derives $w_{i_2}^{j_2 + 2 \delta}$.
\label{thm:correct}
\end{theorem}
\proof{
Fix $i$ and $j$.

\begin{figure*}
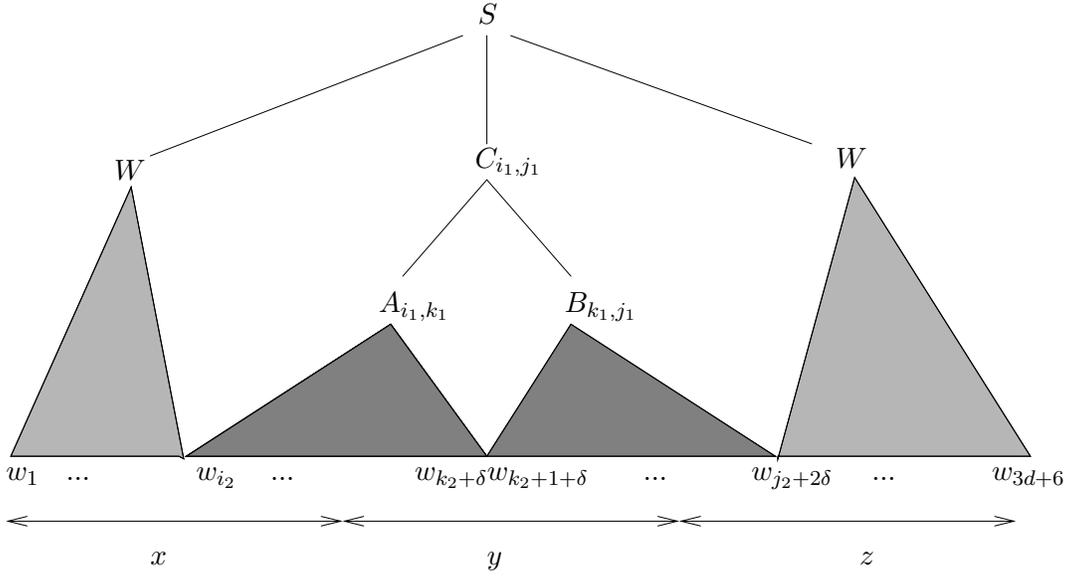

\begin{center}
\include{derive}
\end{center}
\caption{Schematic of the derivation process when $a_{ik} = b_{kj} =
1$.  The substrings derived by
$A_{i_1,k_1}$ and $B_{k_1,j_1}$ lie right next to each other.}
\label{fig:schematic}
\end{figure*}

Let us prove the ``only if'' direction first.
Thus, suppose  $c_{ij} = 1$.  Then there exists a
$k$ such that $a_{ik} = b_{kj} = 1$.  Figure \ref{fig:schematic} sketches how $C_{i_1,j_1}$ c-derives $w_{i_2}^{j_2 + 2 \delta}$.
\begin{claim}
$C_{i_1, j_1} \r w_{i_2}^{j_2 + 2 \delta}$.
\label{claim1}
\end{claim}
The production
$C_{i_1, j_1} \longrightarrow A_{i_1,k_1} B_{k_1,j_1}$ is
one of the $C$-rules in our grammar.
Since $a_{ik} = 1$,
$A_{i_1, k_1} \longrightarrow w_{i_2} \W w_{k_2 + \delta}$
is one of our $A$-rules, and since $b_{kj} = 1$,
$B_{k_1, j_1} \longrightarrow w_{k_2 + 1 + \delta} \W w_{j_2 + 2 \delta}$
is one of our $B$-rules.
Finally, since $i_2 + 1 < (k_2 + \delta) - 1 $ and 
$(k_2 + 1 + \delta) + 1 \leq (j_2 + 2 \delta) - 1$, we have
$\W \r w_{i_2 + 1}^{k_2 + \delta - 1}$ and
$\W \r w_{k_2 + 2 + \delta}^{j_2 + 2 \delta - 1}$,
since both substrings are of length at least one.
Therefore,
\begin{eqnarray*}
C_{i_1, j_1} &\Rightarrow{}{}{}{}{}{}{}&A _{i_1, k_1} B_{k_1, j_1} \\
 & \r &
\underbrace{w_{i_2} \W w_{k_2 + \delta}}_{\mbox{derived by } A_{i_1, k_1}}
\underbrace{ w_{k_2 + 1 + \delta} \W w_{j_2 + 2 \delta}}_{\mbox{derived
by } B_{k_1, j_1} } \\
 &\r& w_{i_2}^{j_2 + 2 \delta}. 
\end{eqnarray*}
\hfill \fbox{\hspace{.01pt}} 

\begin{claim}
$S \r w_1^{i_2-1}C_{i_1,j_1}  w_{j_2 + 2 \delta + 1}^{3\lo +6}.$
\label{claim2}
\end{claim}
This claim is essentially trivial, since 
by the definition of the $S$-rules, we know that $S \r \W C_{i_1,
j_1} \W$. 
We need only show that neither $w_1^{i_2 - 1}$ nor $w_{j_2 + 2
\delta + 1}^{3\lo + 6}$ is the empty string (and hence can be derived by $\W$); since $1 \leq
i_2 - 1$ and $j_2 + 2 \delta + 1 \leq 3\lo + 6$, the claim holds.
\hfill \fbox{\hspace{.01pt}} 

\medskip

Claims~\ref{claim1} and \ref{claim2} together
prove that $C_{i_1, j_1}$ c-derives $w_{i_2}^{j_2 + 2
\delta}$, as required.\footnote{This proof would have been simpler if we had
allowed $\W$ to derive the empty string.  However, we avoid epsilon-productions in order to facilitate the conversion to Chomsky normal
form discussed in the next section.}

Next we prove the ``if'' direction.
Suppose $C_{i_1, j_1}$ c-derives $w_{i_2}^{j_2 + 2 \delta}$,
which by definition means 
$C_{i_1, j_1} \r w_{i_2}^{j_2 + 2 \delta}$.
This can only arise through the application of a $C$-rule:
$$C_{i_1, j_1} \Rightarrow A_{i_1, k'} B_{k', j_1}
\r w_{i_2}^{j_2 + 2 \delta}$$
for some $k'$.
It must be the case that for some $\ell$,
$A_{i_1, k'} \r w_{i_2}^\ell$ and 
$B_{k', j_1} \r w_{\ell+1}^{j_2 + 2 \delta}$.
But then we must have
the productions
$A_{i_1, k'} \longrightarrow w_{i_2} \W w_{\ell}$
and $B_{k', j_1} \longrightarrow w_{\ell + 1} \W w_{j_2 + 2 \delta}$
with $\ell = k'' + \delta$ for some $k''$.
But we can only have such productions if there exists a number $k$
such that $k_1 = k'$, $k_2 = k''$,
$a_{ik} = 1$, and $b_{kj} = 1$; and this implies
that $c_{ij}=1$.
} 

Examination of the proof reveals that we also have the following
two corollaries.

\begin{corollary}
For $1 \leq i,j \leq m$, $c_{ij} = 1$ if and only if $C_{i_1,j_1} \r
w_{i_2}^{j_2 + 2 \delta}$.  Hence, c-derivation and derivation are
equivalent for the $C_{p,q}$ nonterminals.
\label{derive}
\end{corollary}

\begin{corollary}
$S \r w$ if and only if $C$ is not the all-zeroes matrix.
\end{corollary}

Let us now calculate the size of $G$.  $V$ consists of roughly
$3((\lo^2)^2) \approx m^{4/3}$ nonterminals.  $R$ contains about
$6\lo$ $W$-rules and $(\lo^2)^2 \approx m^{4/3}$ $S$-rules.  There are
at most $m^2$ $A$-rules, since we have $A$-rules only for each non-zero
entry in $A$; similarly, there are at most $m^2$ $B$-rules.  And
lastly, there are $(\lo^2)^3 \approx m^2$ $C$-rules.  Therefore, our
grammar is of size $O(m^2)$ with a very small constant factor;
considering that $G$ encodes $m \times m$ matrices $A$ and $B$, it is
not possible to shrink this much further.

\subsection{Chomsky normal form}
\label{CNF}

We would like our results to cover as large a class of parsers as
possible.  Some parsers, such as CKY, require the input grammar to be
in Chomsky normal form (CNF), that is, where the right-hand side of
every production consists of either exactly two nonterminals or
exactly a single terminal. We therefore wish to construct a CNF
version $G'$ of $G$.  However, not only do we want Theorem
\ref{thm:correct} to hold for $G'$ as well as $G$, but, in order to
preserve time bounds, we also desire that $\ab{G'} = O(|G|)$.

Unfortunately, the standard algorithm for converting CFGs to CNF can
yield a quadratic blow-up in the number of productions in the grammar
\cite{Hopcroft:79a} and thus is clearly unsatisfactory for our
purposes.  However, since $G$ contains no epsilon-productions or unit
productions, it is easy to convert $G$ by adding a small number of
record-keeping nonterminals and productions, with the resultant
grammar $G'$ having very similar parse trees --- in particular, the
set of substrings that are c-derived by the $C_{p,q}$ nonterminals are
the same in each grammar.  Figure \ref{fig:cnf} gives the productions
of $G'$.  Note that $G'$ has only $O(\lo)$ more productions and
nonterminals, and so $\ab{G'} = O(m^2)$ as well.

\begin{figure}
\begin{center}
\begin{tabular}{lcll}
$W$ & $\longrightarrow$ & $W_{\ell} W | w_{\ell}$  & $(1 \leq \ell \leq 3\lo +
6)$ \\
$W_{\ell}$ & $\longrightarrow$ & $w_{\ell}$ & $(1 \leq \ell \leq 3\lo +
6)$ \\
\\
$A_{i_1,j_1}$ & $\longrightarrow$ &  $W_{i_2} X_{j_2 + \delta}$ & (one
for each nonzero entry $a_{ij}$ in $A$) \\
$X_{j_2 + \delta}$ & $\longrightarrow$ & $W W_{j_2 + \delta}$ & $(2
\leq j_2 \leq \lo+1)$ \\ 
\\
$B_{i_1,j_1}$  &  $\longrightarrow$ &  $W_{i_2+1+\delta} X_{j_2 +
2\delta}$ & (one
for each nonzero entry $b_{ij}$ in $B$)\\
$X_{j_2 + 2\delta}$ & $\longrightarrow$ & $W W_{j_2 + 2\delta}$ & $(2
\leq j_2 \leq \lo+1)$ \\ 
\\
${C_{p,q}}$ &  $\longrightarrow$ &  ${A_{p,r} B_{r,q}}$ & $ (1 \leq p,
q, r \leq \lo^2)$ \\
\\
$S$ & $\longrightarrow$ & $W  T$ &  \\
$T$ & $\longrightarrow$ & $C_{p,q} W$ & ($1 \leq p,q \leq \lo^2$) 
\end{tabular}
\end{center}
\caption{\label{fig:cnf} A Chomsky normal form version of the
productions of the grammar from the previous section.}
\end{figure}

\subsection{Time bounds}
\label{sec:timebounds}

We are now in a position to show the relation between time bounds for
Boolean matrix multiplication and time bounds for CFG parsing.

\begin{theorem} Any c-parser $P$ with running time $O(\tg{g} \tw{\slength})$
on grammars of size $g$ and strings of length $\slength$ can be converted
into a BMM algorithm $M_P$ that runs in time $O( \max ( m^2, \tg{m^2} 
\tw{m^{1/3}}))$.  In particular, if $P$ takes time $O(g\slength^{3
-\myeps})$, then $M_P$ runs in time $O(m^{3 - \myeps / 3})$.
\label{time}
\end{theorem}
\proof{$M_P$ acts as sketched in Figure \ref{fig:createBMM}.  More
precisely, given two Boolean $m \times m$ matrices $A$ and $B$, it
constructs $\GCNF$ (or $G'$, as required) and $w$ as described above.
It feeds $\GCNF$ and $w$ to $P$, which outputs oracle $\output$.  To
compute the product matrix $C$, $M_P$ requests from the oracle the
value of $\output(C_{i_1,j_1},i_2, j_2+2\delta)$ (that is, whether or
not $C_{i_1, j_1}$ derives or c-derives\footnote{By corollary
\ref{derive}, the two notions are equivalent in this case.} $w_{i_2}^{j_2 + 2 \delta}$) for
each $i$ and $j$, $ 1 \leq i,j \leq m$, setting $c_{ij}$ to one if and
only if the answer is ``yes''. 

The running time of $M_P$ is computed as follows.  It takes $O(m^2)$
time to read the two input matrices.  Since $\GCNF$ is of size $O(m^2)$
and $\ab{w} = O(m^{1/3})$, it takes $O(m^2)$ time to build the input
to $P$, which then computes $\output$ in time $O(\tg{m^2}
\tw{m^{1/3}})$.  Retrieving $C$ takes $O(m^2)$ since, by definition of
c-parser, each query to the oracle takes constant time.  So the total
time spent by $M_P$ is $O(\max (m^2, \tg{m^2} \tw{m^{1/3}}))$, as 
claimed.

Note that if we redefine c-parsing so that oracle queries take
$f(g,\slength)$ time instead of constant time, where $g$ is the size
of the grammar and $\slength$ is the length of the string, then the
bound changes to $O(\max (m^2f(g,\slength), \tg{m^2} \tw{m^{1/3}}))$;
as long as $f$ is polylogarithmic, the second argument of the maximum
in the bound surely dominates.

In the case where $\tg{g} = g$ and $\tw{\slength} = \slength^{3
-\myeps}$, $M_P$ has a running time of $O(m^2 (m^{1/3})^{3 - \myeps})
= O(m^{3 -\myeps /3})$.  } 

The case in which $P$ takes time linear in the grammar size is of the
most interest, since, as mentioned above, in natural language
processing applications the grammar tends to be far larger than the
strings to be parsed.  In this case, our result directly converts
any improvement in the exponent for CFG parsing to a reduction in the
exponent for BMM.  For example, observe that Theorem \ref{time}
translates the running time of the standard CFG parsers,
$O(g\slength^3)$, into the running time of the standard BMM algorithm,
$O(m^3)$.  Also, a c-parser with running time $O(g\slength^{2.43})$
would yield a matrix multiplication algorithm rivalling that of
Strassen's \shortcite{Strassen:mult}, and a c-parser with running time
better than $O(g\slength^{1.12})$ could be converted into a BMM method
faster than \namecite{Coppersmith+Winograd:90a}.  As per the
discussion above, even if such parsers exist, they would in all
likelihood not be very practical.

\subsubsection{Parameter choices}

Since \namecite{Valiant:cfl} proved that an $O(m^{3 - \myeps})$ BMM
algorithm can be transformed into a parser with time complexity
$O(\slength^{3-\myeps})$ in the string length,
it is natural to ask whether our technique could yield the stronger
result (if it is in fact true) that a CFG parser running in time
$O(g\slength^{3-\myeps})$ can be converted into an $O(m^{3 - \myeps})$
BMM algorithm.  We now explain why such a result cannot be obtained by
a straightforward modification of the reduction method we described
above.

Our run-time results are based on a particular choice of where to
divide matrix indices into ``high order bits'' and ``low order bits'';
in particular, we set $\lo$, which parametrizes the number of low
order bits, to $\lo = \lceil m^{1/3} \rceil$.  We determined this
value by considering the effect of $\lo$ on the size of the resulting
grammar and string:  roughly speaking, a larger value shrinks the
former but expands the latter.  For convenience, let us set $\lo
= m^\ell$, and consider how to pick $\ell$.

Since combining the higher-order bits and the lower-order bits yields
a matrix index of magnitude at most $m$, it follows that the string
has size $O(m^\ell)$ and the grammar will have size $O(m^2 +
(m^{1-\ell})^3)$ (the first term comes from the inclusion of the $A$-
and $B$-rules, and the second term comes from the fact that the
$C$-rules have to include the higher-order bits for three matrix
indices).  Hence, a parser with run-time complexity
$O(g\slength^{3-\myeps})$ yields a BMM algorithm with run-time
complexity $O(m ^{2 + (3 - \myeps)\ell} + m^{3 - \myeps \ell })$.
Inspection reveals that when $\ell > 1/3$, the first term dominates;
when $\ell < 1/3$, the second term dominates; and the lowest upper
bound occurs at the ``crossing point'' where $\ell = 1/3$.


\section{Related results}

We have shown that the existence of a fast practical CFG parsing
algorithm would yield a fast practical BMM algorithm. Given that fast
practical BMM algorithms are thought not to exist, this establishes a
limitation on the efficiency of practical CFG parsing, and helps
explain why there has been very little success in developing practical
sub-cubic general CFG parsers.

There have been a number of related results regarding the time
complexity of context-free grammar parsing and the relationship
between this and other problems.  We survey these results below.

As mentioned above, the asymptotically fastest (although not
practical) general context-free parsing algorithm is due to
\namecite{Valiant:cfl}, who showed that the problem can be reduced to
Boolean matrix multiplication (this is the ``opposite direction'' of
the reduction we present).  His algorithm shows that the worst-case
dependence of the speed of CFG parsing on the input string length is
$O(M(\slength))$, where $M(m)$ is the time it takes to multiply two $m
\times m$ Boolean matrices together.  (\namecite{Rytter:95a} provides
an alternate version of this algorithm with the same asymptotic
complexity.)

Methods for reducing Boolean matrix multiplication to context-free
grammar parsing were previously considered by \namecite{Ruzzo:79a}.
He proved that the problem of producing all possible parses of a
string of length $\slength$ with respect to a context-free grammar is
at least as hard as multiplying two $\sqrt{\slength} \times
\sqrt{\slength}$ Boolean matrices together. His technique encodes most of the information
about the matrices in  strings (as opposed to in the grammar, as in our
method). Ruzzo's result does not serve to explain why practical
sub-cubic CFG parsing algorithms have been so difficult to produce,
since using his reduction translates even a parser running in time
proportional to $\slength^{1.5}$ to a cubic-time BMM algorithm.

Harrison and Havel \cite{Harrison+Havel:74,Harrison:78a} note that
there is a reduction of $m \times m$ BMM {\em checking} to
context-free {\em recognition} (a BMM checker takes as input three Boolean
matrices $A$, $B$, and $C$ and reveals whether or not $C$ is the
Boolean product of $A$ and $B$).  These two decision problems are
clearly related to the algorithmic problems we consider in this
paper. However, this reduction, like Ruzzo's, also converts a parser
running in time proportional to $\slength^{1.5}$ to a cubic-time BMM
checking algorithm, which, again, is not as strong a result as ours.

The problem of {\em on-line} CFL recognition is to proceed through
each prefix $w_1^i$ of the input string $w$, determining whether or
not $w_1^i$ is generated by the input context-free grammar before
reading the next ($(i+1)$th) input symbol.  The study of the
complexity of this problem has a long history; in fact, the landmark
paper of \namecite{Hartmanis+Stearns:65a} that introduced the notions
of time and space complexity contains an example of a CFL for which
on-line recognition of strings of length $\slength$ takes more than
$\slength$ steps.  Currently, the best known lower bound for this problem is
$\Omega(\frac{\slength^2}{\log \slength})$  \cite{Seiferas:lower,Gallaire}.
However, on-line recognition is a more difficult task than the standard
CFL recognition problem (indeed, it is the extra constraints imposed
by the on-line requirement that make it easier to
prove lower bounds), and so these results do not translate to the
usual recognition paradigm.  To date, there are no non-trivial lower
bounds known for general CFL recognition.

Relationships between parsing other grammatical formalisms and
multiplying Boolean matrices have also been explored.  In particular,
several researchers have looked at {\em Tree Adjoining Grammar} (TAG)
\cite{Joshi+Levy+Takahashi:75a}, an elegant formalism based on
modifying tree structures.  TAGs have strictly greater generative
capacity than context-free grammars, but at the price of being
(apparently) harder to parse: standard algorithms run in time
proportional to $\slength^6$, although
\namecite{Rajasekaran+Yooseph:95a} adapt Valiant's
\shortcite{Valiant:cfl} technique to get an asymptotically faster
parser using BMM.  \namecite{Satta:bmm} gives a reduction of Boolean
matrix multiplication to tree-adjoining grammar parsing, demonstrating
that any substantial improvement over $O(g\slength^6)$ for TAG parsing
would result in a sub-cubic BMM algorithm. Our reduction was inspired
by Satta's and resembles his in the way that matrix information is
encoded in a grammar.  However, Satta's reduction explicitly relies on
TAG properties that allow non-context-free languages to be generated,
and so cannot be directly applied to CFG parsing.

\section{Acknowledgments}

Thanks to Zvi Galil, Joshua Goodman, Rebecca Hwa, Jon Kleinberg, Giorgio Satta,
Stuart Shieber, Les Valiant, and the anonymous referees for many
helpful comments and conversations.  A preliminary conference
version of this paper appeared in the {\em Proceedings of the 35th
Annual Meeting of the Association for Computational Linguistics}, pp. 9--15;
thanks to those reviewers for their comments and suggestions. This
material is based upon work supported in part by the National Science
Foundation under Grant No. IRI-9350192, an NSF Graduate Fellowship,
and an AT\&T GRPW/ALFP grant.  Any opinions, findings, and conclusions
or recommendations expressed above are those of the author and do not
necessarily reflect the views of the National Science Foundation.

\bibliography{master,../mm}

\end{document}

%% file: derive.tex
\setlength{\unitlength}{0.00083333in}
\begingroup\makeatletter\ifx\SetFigFont\undefined
\def\x#1#2#3#4#5#6#7\relax{\def\x{#1#2#3#4#5#6}}%
\expandafter\x\fmtname xxxxxx\relax \def\y{splain}%
\ifx\x\y   
\gdef\SetFigFont#1#2#3{}%
\else
\gdef\SetFigFont#1#2#3{\begingroup
  \count@#1\relax \ifnum 25<\count@\count@25\fi
  \def\x{\endgroup\@setsize\SetFigFont{#2pt}}%
  \expandafter\x
    \csname \romannumeral\the\count@ pt\expandafter\endcsname
    \csname @\romannumeral\the\count@ pt\endcsname
  \csname #3\endcsname}%
\fi
\gdef\SetFigFont#1#2#3{}
\fi\endgroup
\begin{picture}(6673,3525)(0,-10)
\put(2945,3402){\makebox(0,0)[lb]{\smash{{{\SetFigFont{12}{14.4}{rm}$S$}}}}}
\path(2475,1827)(3000,2427)(3525,1827)
\put(2925,2502){\makebox(0,0)[lb]{\smash{{{\SetFigFont{12}{14.4}{rm}$C_{i_1,j_1}$}}}}}
\put(2325,1602){\makebox(0,0)[lb]{\smash{{{\SetFigFont{12}{14.4}{rm}$A_{i_1,k_1}$}}}}}
\put(3485,1602){\makebox(0,0)[lb]{\smash{{{\SetFigFont{12}{14.4}{rm}$B_{k_1,j_1}$}}}}}
\path(12,297)(2088,297)
\path(132.000,327.000)(12.000,297.000)(132.000,267.000)
\path(1968.000,267.000)(2088.000,297.000)(1968.000,327.000)
\path(2112,297)(4188,297)
\path(2232.000,327.000)(2112.000,297.000)(2232.000,267.000)
\path(4068.000,267.000)(4188.000,297.000)(4068.000,327.000)
\path(4212,297)(6288,297)
\path(4332.000,327.000)(4212.000,297.000)(4332.000,267.000)
\path(6168.000,267.000)(6288.000,297.000)(6168.000,327.000)
\put(900,27){\makebox(0,0)[lb]{\smash{{{\SetFigFont{12}{14.4}{rm}$x$}}}}}
\put(3000,27){\makebox(0,0)[lb]{\smash{{{\SetFigFont{12}{14.4}{rm}$y$}}}}}
\put(5325,27){\makebox(0,0)[lb]{\smash{{{\SetFigFont{12}{14.4}{rm}$z$}}}}}
\path(3000,3327)(3000,2652)
\path(900,2577)(2850,3327)
\texture{44555555 55aaaaaa aa555555 55aaaaaa aa555555 55aaaaaa aa555555 55aaaaaa 
	aa555555 55aaaaaa aa555555 55aaaaaa aa555555 55aaaaaa aa555555 55aaaaaa 
	aa555555 55aaaaaa aa555555 55aaaaaa aa555555 55aaaaaa aa555555 55aaaaaa 
	aa555555 55aaaaaa aa555555 55aaaaaa aa555555 55aaaaaa aa555555 55aaaaaa }

\shade\path(3000,702)(2400,1527)(1125,702)(3000,702)
\path(3000,702)(2400,1527)(1125,702)(3000,702)

\shade\path(3000,702)(3525,1527)(4800,702)(3000,702)
\path(3000,702)(3525,1527)(4800,702)(3000,702)
\path(5025,2652)(3150,3327)
\shade\path(3000,702)(2400,1527)(1125,702)(3000,702)
\path(3000,702)(2400,1527)(1125,702)(3000,702)
\texture{88555555 55000000 555555 55000000 555555 55000000 555555 55000000 
      555555 55000000 555555 55000000 555555 55000000 555555 55000000 
      555555 55000000 555555 55000000 555555 55000000 555555 55000000 
      555555 55000000 555555 55000000 555555 55000000 555555 55000000 }

\shade\path(1080,702)(30,702)(780,2382)
        (1110,687)(1080,702)
\path(1080,702)(30,702)(780,2382)
        (1110,687)(1080,702)

\shade\path(4815,702)(6390,702)(5295,2442)
      (4815,687)(4815,702)(4815,702)
\path(4815,702)(6390,702)(5295,2442)
      (4815,687)(4815,702)(4815,702)

\put(0,552){\makebox(0,0)[lb]{\smash{{{\SetFigFont{12}{14.4}{rm}$w_1$}}}}}
\put(3000,552){\makebox(0,0)[lb]{\smash{{{\SetFigFont{12}{14.4}{rm}$w_{k_2+1+\delta}$}}}}}
\put(2550,552){\makebox(0,0)[lb]{\smash{{{\SetFigFont{12}{14.4}{rm}$w_{k_2+\delta}$}}}}}
\put(4650,552){\makebox(0,0)[lb]{\smash{{{\SetFigFont{12}{14.4}{rm}$w_{j_2
+ 2\delta}$}}}}}
\put(1185,552){\makebox(0,0)[lb]{\smash{{{\SetFigFont{12}{14.4}{rm}$w_{i_2}$}}}}}
\put(675,2427){\makebox(0,0)[lb]{\smash{{{\SetFigFont{12}{14.4}{rm}$W$}}}}}
\put(1650,552){\makebox(0,0)[lb]{\smash{{{\SetFigFont{34}{40.8}{rm}...}}}}}
\put(375,552){\makebox(0,0)[lb]{\smash{{{\SetFigFont{34}{40.8}{rm}...}}}}}
\put(3975,552){\makebox(0,0)[lb]{\smash{{{\SetFigFont{34}{40.8}{rm}...}}}}}
\put(5400,552){\makebox(0,0)[lb]{\smash{{{\SetFigFont{34}{40.8}{rm}...}}}}}
\put(5175,2502){\makebox(0,0)[lb]{\smash{{{\SetFigFont{12}{14.4}{rm}$W$}}}}}
\put(6150,552){\makebox(0,0)[lb]{\smash{{{\SetFigFont{12}{14.4}{rm}$w_{3\lo+6}$}}}}}
\end{picture}